\documentclass[journal]{IEEEtran}

\usepackage{amsmath}
\usepackage{amssymb}
\usepackage{bm}
\usepackage{graphicx}
\usepackage{booktabs}
\usepackage{multirow}
\usepackage{pifont}
\usepackage[table]{xcolor}
\usepackage{cite}
\usepackage{url}
\usepackage[hidelinks]{hyperref}

\newcommand{\cmark}{\textcolor{green!60!black}{\ding{51}}}
\newcommand{\xmark}{\textcolor{red!70!black}{\ding{55}}}
\newcommand{\tmark}{$\triangle$}

\begin{document}

\title{Fog-FWL: A Large-Scale Full-Waveform LiDAR Dataset for Recovering Fog-Hidden Objects}

\author{Kazuma Ikeda$^{1,*}$, Kotaro Oishi$^{1,*}$, Ryosei Hara$^{1}$, Ryo Yoshida$^{1}$, Takahiro Kado$^{2}$, Ibuki Fujioka$^{2}$,\\ Taro Beppu$^{2}$, Mariko Isogawa$^{1}$, and Kentaro Yoshioka$^{1}$%
\thanks{$^{*}$These authors contributed equally to this work.}%
\thanks{$^{1}$Keio University, Japan.}%
\thanks{$^{2}$Sony Semiconductor Solutions, Japan.}}

\maketitle

\begin{abstract}
LiDAR is a core sensor for autonomous driving, yet it is highly vulnerable to fog. Fog scattering produces strong backscatter near the sensor and attenuates reflections from objects behind it, so weak object returns are often discarded by the peak detection that converts received signals into point clouds. Full-waveform LiDAR (FWL), which records the received optical signal before peak detection, retains these weak returns, but the lack of suitable data has prevented research on recovering fog-hidden objects from FWL. We present \textbf{Fog-FWL}, a dataset foundation for this task that consists of a real-world benchmark, \textbf{Fog-FWL-Real}, and a large-scale synthetic training set, \textbf{Fog-FWL-Sim}. Fog-FWL-Real contains 7,527 real foggy FWL frames from 8 scenes and 58 sequences, which is $579\times$ larger than the only existing real foggy FWL data, together with histogram-level \textit{Object}/\textit{Fog}/\textit{Noise} labels, 10,719 3D bounding boxes, and three difficulty levels defined by the attenuation of object reflections. Fog-FWL-Sim contains 12,000 frames generated by our physics-based simulator, Fog-FWL-Simulator, which integrates spatially heterogeneous fog density fields, time-resolved light transport including multiple scattering, and a photon-counting SPAD receiver model. The two subsets share the same sensor format and annotation protocol, enabling a strict sim-to-real evaluation in which no real waveforms are used for training. Our analysis shows that, under the hardest condition, object returns are included in the three strongest peaks for only 34.9\% of object rays, underscoring the need for waveform-level recovery. The dataset will be released upon acceptance.
\end{abstract}


\section{Introduction}
\label{sec:intro}

Autonomous driving technologies are being actively developed to reduce traffic accidents and improve mobility, and commercial robotaxi services already operate in several cities in the United States and China~\cite{waymo, apollo}. However, these deployments are concentrated in warm, clear-weather cities, while deployment in regions prone to rain, fog, and snow remains limited~\cite{waymo2025allweather, clickondetroit2026waymo}. Among adverse weather conditions, fog severely impairs visibility through the scattering and attenuation of light by suspended water droplets and contributes to approximately 40,000 traffic crashes annually in the United States alone~\cite{fhwa_lowvisibility}. Most commercially operated autonomous vehicles rely on Light Detection and Ranging (LiDAR) to acquire accurate 3D geometry independent of ambient lighting, yet LiDAR remains highly susceptible to fog.

In foggy environments, LiDAR measurements contain backscattered signals from fog particles, while reflections from objects are attenuated as light propagates through the medium~\cite{rasshofer2011influences, bijelic2018benchmark, bijelic2020seeing}. Real-world fog is also spatially heterogeneous; when a localized dense fog bank lies between the sensor and an object, the weak object reflection can be buried in scattering noise, and vehicles or pedestrians may go undetected.

Many methods restore object geometry from degraded point clouds through denoising and completion~\cite{kaito2026enhancing, seppanen2024self, man2021multi, sun2026weather}. However, conventional LiDARs generate point clouds by applying peak detection to the received signals, and attenuated object reflections that fall below the strongest returns are discarded at this stage. Post-processing on point clouds can only exploit the returns that survive peak detection, which limits recovery when most object returns are lost.

Full-waveform LiDAR (FWL) data, i.e., the raw received optical waveforms before point cloud conversion, offers a way around this limitation. Because FWL records the temporal profile of the received intensity, it retains weak reflections and overlapping return components that are lost in point clouds. Recent studies have exploited FWL for object recognition and noise removal under complex reflection and scattering conditions~\cite{scheuble2025lidar, ikeda2026ghost, oishi2026fwltopm}. Progress on recovering fog-hidden objects from FWL, however, is held back by data. Real foggy FWL data with annotations are scarce: the only existing real foggy FWL data consist of 13 frames~\cite{scheuble2025lidar}, which is insufficient for evaluation stratified by fog severity and is not publicly available. Existing foggy FWL simulation~\cite{scheuble2025lidar} superimposes a spatially uniform analytical fog model onto clear-weather waveforms, which does not reproduce heterogeneous fog distributions or multiple scattering, and approximates device-specific single-photon avalanche diode (SPAD) responses such as dead time and pile-up only coarsely.

To address these gaps, we present \textbf{Fog-FWL}, an FWL dataset foundation for recovering objects hidden by fog. Fog-FWL comprises a real-world benchmark, \textbf{Fog-FWL-Real}, and a large-scale training set, \textbf{Fog-FWL-Sim}, generated by our physics-based simulator, \textbf{Fog-FWL-Simulator}. Fog-FWL-Real is intended solely for evaluation, and Fog-FWL-Sim for training, so that methods can be evaluated in a strict sim-to-real setting without using any real measured waveforms for training. The dataset will be released upon acceptance.

Our contributions are summarized as follows:
\begin{itemize}
\item We construct \textbf{Fog-FWL-Real}, the first large-scale real-world foggy FWL dataset, with 7,527 frames from 8 scenes and 58 sequences ($579\times$ larger than the only existing real foggy FWL data~\cite{scheuble2025lidar}). It provides histogram-level \textit{Object}, \textit{Fog}, and \textit{Noise} labels, 10,719 3D bounding boxes, and three difficulty levels defined by the attenuation of object reflections.
\item We develop \textbf{Fog-FWL-Simulator}, which integrates diverse CARLA driving scenes, heterogeneous fog density fields, time-resolved volumetric light transport rendering with Mitsuba~3, and a photon-counting SPAD receiver model, and use it to build \textbf{Fog-FWL-Sim}, a training set of 12,000 frames with the same sensor format and annotation protocol as Fog-FWL-Real.
\item We analyze the dataset and show that, under the Hard condition, object returns appear among the three strongest peaks for only 34.9\% of object rays, quantifying the information lost by peak detection in dense fog.
\end{itemize}

\section{Related Work}
\label{sec:related}

\subsection{LiDAR Perception in Fog}
Fog degrades LiDAR point clouds in two ways: spurious points are added by backscatter from fog particles, and object points are lost as reflections are attenuated. For the former, many methods identify and remove weather-induced points; Seppanen et al.~\cite{seppanen2024self} and Kaito et al.~\cite{kaito2026enhancing} select object returns among multiple echoes to remove noise in snow and fog, respectively. For the latter, Man et al.~\cite{man2021multi} integrate multiple returns of multi-echo LiDAR to improve detection, and Sun et al.~\cite{sun2026weather} restore geometry from weather-degraded point clouds. All of these methods take point clouds or echoes after peak detection as input. Because typical LiDARs output only the two or three strongest peaks per ray, object reflections weaker than fog backscatter are often absent from the input under dense fog. Fusion with cameras or radar~\cite{bijelic2020seeing, yoon2025equirectangular, chae2025doppler} is a complementary direction but requires additional sensors and inter-sensor calibration.

\subsection{Full-Waveform LiDAR Datasets}
Recovering weak object reflections in fog requires a dataset that provides (a)~real fog scattering, (b)~raw FWL waveforms before peak detection, (c)~fine-grained labels separating object reflections from backscatter, (d)~3D bounding boxes to assess the effect on downstream recognition, and (e)~a difficulty protocol based on the attenuation of object reflections, with sufficient data at each level. Large-scale driving benchmarks~\cite{caesar2020nuscenes, sun2020waymo, geiger2013kitti, chang2019argoverse, wilson2023argoverse} and adverse-weather datasets~\cite{bijelic2020seeing, xiao20233d, paek2022k, sheeny2021radiate, carballo2020libre} provide rich point clouds and images but no raw waveforms. PixSet~\cite{deziel2021pixset} provides FWL waveforms with 3D detection labels, and Ghost-FWL~\cite{ikeda2026ghost} provides peak-level labels, but both are recorded in clear weather. Scheuble et al.~\cite{scheuble2025lidar} were the first to study real foggy FWL data, but their real data consist of 13 frames, do not include detection labels, and are not publicly available. Table~\ref{tab:dataset_comparison} summarizes these datasets.

\begin{table}[t]
\centering
\caption{Comparison of real-world LiDAR datasets for fog.}
\label{tab:dataset_comparison}
\scriptsize
\setlength{\tabcolsep}{2pt}
\renewcommand{\arraystretch}{1.15}
\begin{tabular*}{\columnwidth}{@{\extracolsep{\fill}} l c c c c c c c c @{}}
\toprule
Dataset & Year & Pub. & Fog & FWL & \shortstack{Obj.\\Det.} & \shortstack{Peak\\label} & \shortstack{Diff.\\Eval.} & \shortstack{Fog FWL\\Frames} \\
\midrule
STF Dataset~\cite{bijelic2020seeing} & 2020 & \cmark & \cmark & \xmark & \cmark & \xmark & \xmark & -- \\
Leddar PixSet~\cite{deziel2021pixset} & 2021 & \cmark & \xmark & \cmark & \cmark & \xmark & \xmark & 0 \\
Scheuble et al.~\cite{scheuble2025lidar} & 2025 & \xmark & \cmark & \cmark & \xmark & \cmark & \tmark & 13 \\
Ghost-FWL~\cite{ikeda2026ghost} & 2026 & \cmark & \xmark & \cmark & \xmark & \cmark & \xmark & 0 \\
\midrule
\textbf{Fog-FWL-Real (Ours)} & \textbf{2026} & \cmark & \cmark & \cmark & \cmark & \cmark & \cmark & \textbf{7.5k} \\
\bottomrule
\end{tabular*}
\vspace{2pt}

\raggedright\scriptsize Obj. Det.: object detection labels. Peak label: peak-level labels. Diff. Eval.: difficulty levels based on object signal attenuation. Fog-FWL will be released upon acceptance.
\end{table}

\subsection{FWL Simulation}
Because real foggy FWL data are costly to collect and annotate, training on synthetic data and transferring to real data is a practical option. General LiDAR simulators such as CARLA~\cite{dosovitskiy2017carla} and Isaac Sim~\cite{gao2026nvidia} generate point clouds by ray casting and cannot produce FWL waveforms containing temporally spread fog responses. Hahner et al.~\cite{hahner2021fog} applied an analytical model of homogeneous fog to clear-weather point clouds. Scheuble et al.~\cite{scheuble2025lidar} extended this idea to FWL by adding exponential attenuation, a near-range scattering peak fitted to measured fog waveforms, and a constant term approximating multiple scattering to clear-weather waveforms. These models assume the same fog for every ray and therefore cannot represent ray-dependent attenuation and backscatter caused by fog localized in front of objects, nor light paths in which fog-scattered light is reflected by objects. Frameworks for time-resolved rendering of participating media exist~\cite{jarabo2014framework, royo2025mitransient}, but to our knowledge their application to LiDAR waveform generation has been limited to scenes without participating media~\cite{behari2026denali}. Fog-FWL-Simulator places a heterogeneous fog density field in the 3D scene as a participating medium and combines time-resolved rendering with a SPAD receiver model.

\section{Fog-FWL-Real}
\label{sec:real}

Fog-FWL-Real is a real-world FWL dataset for evaluating the recovery of fog-attenuated object reflections and downstream 3D object detection. It contains FWL waveforms captured in fog, histogram-level \textit{Object}, \textit{Fog}, and \textit{Noise} labels, and 3D bounding boxes, comprising 8 scenes, 58 sequences, and 7,527 frames. The dataset is intended for evaluation only.

\subsection{Sensing Setup}
Commercial LiDARs process the received signals internally and output only point clouds, which makes FWL data before peak detection inaccessible. We therefore acquire time-resolved photon-count histograms directly from the FPGA of an in-house SPAD LiDAR~\cite{ikeda2026ghost}. Each frame consists of 400 pixels in azimuth, 512 pixels in elevation, and 700 time bins, with angular resolutions of $0.1^\circ$ (azimuth) and $0.05^\circ$ (elevation), a field of view of $40^\circ\times25.6^\circ$, and a temporal resolution of 1\,ns.

\subsection{Scenes and Objects}
Measurements were conducted in an outdoor parking lot. Across 8 scenes, we varied the background, such as buildings and trees, and the vehicle arrangement, placing one to three vehicles at 10--20\,m from the sensor. In total, the dataset contains approximately 940 million \textit{Object} labels and 10,719 3D bounding boxes. Table~\ref{tab:stats} lists the number of frames per difficulty level.

\subsection{Fog Generation}
Fog was generated with a fog machine (STAGE EVOLUTION PFOG1500~\cite{fog-machine}). Unlike measurements in a weather chamber with spatially homogeneous fog~\cite{scheuble2025lidar}, we target fog localized in front of objects. Such fog attenuates object reflections by different amounts for each ray and causes partial loss of objects. We therefore generated localized fog between the sensor and the target vehicles and recorded FWL data containing fog backscatter and attenuated object reflections. Fog density was controlled by changing the operating interval of the fog machine.

\subsection{Difficulty Levels}
Because fog does not necessarily cover the entire field of view and may be localized in front of objects, we define difficulty based on the attenuation of rays that hit objects. For each ray passing through a 3D bounding box, we compute the ratio $r = I_{\mathrm{fog}} / I_{\mathrm{clear}}$ between the object peak intensities measured with and without fog, and take the mean over the box as the attenuation ratio of the frame. Since frames within a sequence are strongly correlated, difficulty is assigned per sequence using the median attenuation ratio of 20 evenly spaced frames, and all frames in the sequence share that level. We define Easy as $r \geq 0.70$, Moderate as $0.40 \leq r < 0.70$, and Hard as $r < 0.40$; Hard corresponds to data in which objects tend to become locally invisible.

\subsection{Annotation}
Each time bin is assigned one of \textit{Object}, \textit{Fog}, or \textit{Noise}. Here, \textit{Object} denotes reflections from any object surface, including the ground, walls, and buildings, not only vehicles. Annotation uses a fog-free reference waveform obtained by summing approximately 100 frames measured in the same scene and configuration before fog generation.

\textit{Object} labels are assigned once per ray. The bin $k_{\mathrm{ref}}$ of the maximum peak in the reference waveform is taken as the object surface. In the foggy waveform, the nearest peak within $\pm2$ bins ($\pm0.30$\,m) of $k_{\mathrm{ref}}$ is labeled \textit{Object}; if no peak exists in this range, $k_{\mathrm{ref}}$ itself is labeled. This provides supervision even for object reflections that are attenuated below the strongest peaks or are not detected as peaks at all. Rays without an object surface (sky) are excluded based on the temporal concentration (participation ratio) of the reference waveform. Because this decision uses only the reference waveform, whether a ray is labeled does not depend on the fog state.

\textit{Fog} labels are assigned by interval detection rather than peak detection, because fog backscatter tends to form a shoulder-like profile rather than a distinct peak. The mean count of the waveform beyond the object is used as the background level $\lambda$, and intervals in front of the object in which the count exceeds $\lambda+5\sqrt{\lambda}$ and is at least 3 for five or more consecutive bins are taken as fog candidates. The bin with the maximum count in each candidate is labeled \textit{Fog}. To avoid confusion with object reflections, bins beyond three bins before $k_{\mathrm{ref}}$ are excluded, and labels located 0.15\,m or more below the estimated ground plane are removed. Finally, \textit{Object} and \textit{Fog} labels are extended in time to the full width at half maximum of their peaks, and the remaining bins are labeled \textit{Noise}.

3D bounding boxes were annotated manually on point clouds converted from the FWL data. All annotations were verified by experts in computer vision and LiDAR.

\begin{figure*}[t]
\centering
\includegraphics[width=\textwidth]{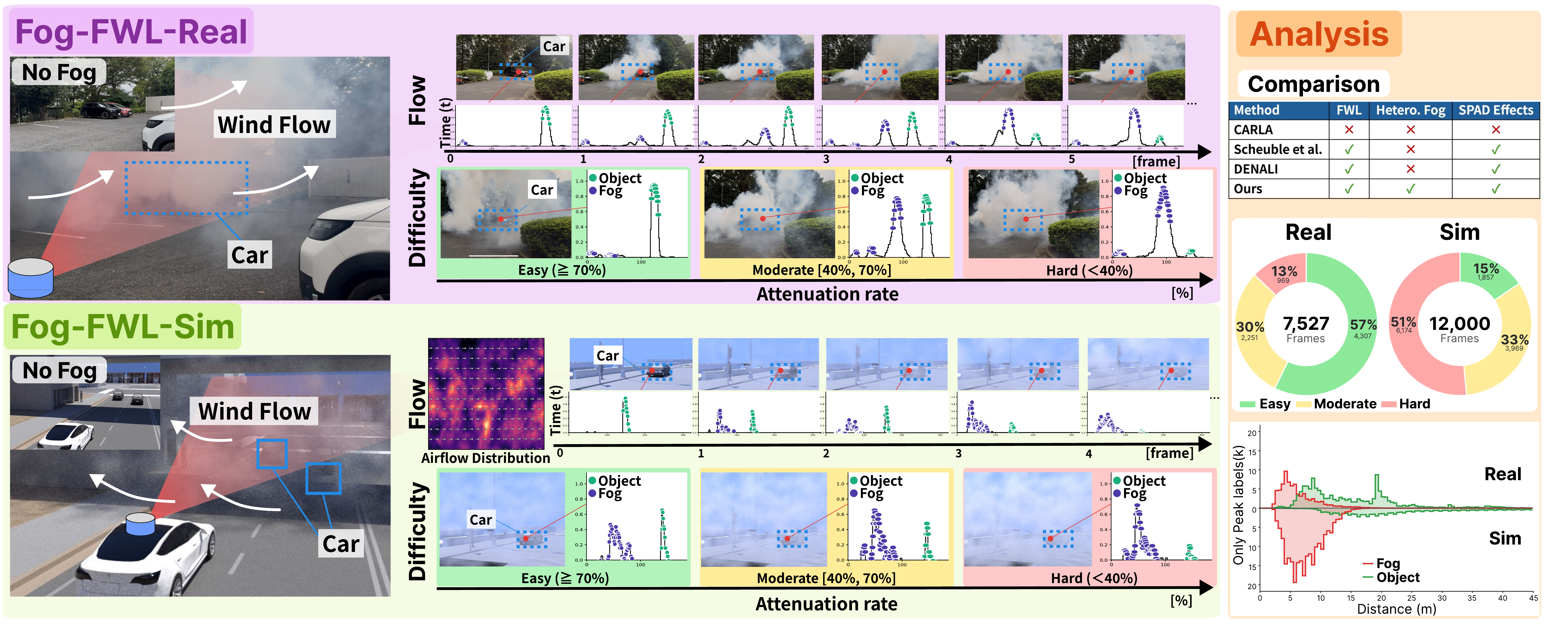}
\caption{Overview of Fog-FWL-Simulator. The pipeline constructs diverse 3D driving scenes in CARLA, generates spatially heterogeneous fog density fields, and performs time-resolved volumetric rendering using Mitsuba~3~\cite{Mitsuba3} and mitransient~\cite{royo2025mitransient} to simulate fog scattering and the attenuation of object reflections. A SPAD receiver model further accounts for photon-counting effects, including saturation, pile-up, and tailing. The resulting waveforms form Fog-FWL-Sim with histogram-level annotations and 3D bounding boxes.}
\label{fig:sim}
\end{figure*}

\section{Fog-FWL-Sim}
\label{sec:sim}

To generate large amounts of FWL data in spatially heterogeneous fog, we build a data generation pipeline, Fog-FWL-Simulator, that integrates 3D scene construction, fog density field generation, time-resolved rendering, and a SPAD receiver model (Fig.~\ref{fig:sim}). For a model trained on synthetic data to transfer to real data, the factors that characterize real waveforms must also appear in synthetic ones. Fog released from a source is advected and diffused and therefore distributed non-uniformly, causing different attenuation and backscatter for each laser ray. In addition, the shape of the received waveform depends not only on fog but also on the response of the SPAD receiver. The pipeline incorporates both factors in three stages. First, vehicles are placed in urban environments of CARLA~\cite{dosovitskiy2017carla} to construct 3D scenes. Second, a localized fog density field, released from a source and advected and diffused, is placed in the scene as a participating medium; the entire scene including fog is rendered in a time-resolved manner with Mitsuba~3~\cite{Mitsuba3} and mitransient~\cite{royo2025mitransient}, and the result is convolved with the emitted pulse shape. Third, the waveform is converted to photon counts by the SPAD receiver model, yielding FWL data in the same format as the real measurements. Because fog-scattered light and object reflections go through the same rendering and detection processes, the attenuation and suppression of object reflections by fog that occur in real hardware also appear in the synthetic waveforms. For each frame, a fog-free waveform with the same scene, viewpoint, and sensor conditions is also generated through the same receiver model and used for annotation and difficulty evaluation.

\subsection{Scene Generation}
We use eight CARLA urban environments (Town01--07 and Town10). The sensor pitch is varied between $-3^\circ$ and $3^\circ$. At each viewpoint, one to three vehicles are placed within $\pm17^\circ$ in azimuth in front of the sensor, at a distance of at least 15\,m with their centers within 25\,m. The scanning configuration is identical to the real sensor in Sec.~\ref{sec:real}: $400\times512\times700$ bins with resolutions of $0.1^\circ$ (azimuth), $0.05^\circ$ (elevation), and 1\,ns, a field of view of $40^\circ\times25.6^\circ$, and the same output format. The light source is a point light at the sensor position, and a single surface bounce is computed. Fog-FWL-Sim contains 12,000 foggy frames from 1,500 viewpoints in each of the eight towns, with one frame per viewpoint; object placement and fog distribution are changed for every viewpoint.

\subsection{Fog Field and Rendering}
To model the heterogeneous density of fog patches localized in front of objects, we define the extinction coefficient $\sigma_t(\mathbf{x})$ at position $\mathbf{x}\in\mathbb{R}^3$ using a plume model that accounts for advection, diffusion, and dilution. Fog is released from a source placed to the side of the line of sight and advected in a direction of 80--88$^\circ$ relative to the line of sight, so that it is distributed mainly in front of the objects. Let $s$ be the downwind distance from the source and $\rho$ the distance from the plume axis. The plume radius $R(s)$ is given by~\cite{morton1956turbulent, pope2000turbulent, taylor1922diffusion, arya1999air}
\begin{equation}
R(s)=r_0+as+b\sqrt{s/U},
\label{eq:plume}
\end{equation}
where $r_0$ is the initial radius, $a$ the linear spreading coefficient, $b$ the turbulent mixing coefficient, and $U$ the advection distance per frame. These parameters and the source position are randomized for each scene. Inside the plume ($0\le s\le L$, $\rho<R(s)$), the extinction coefficient is defined as
\begin{equation}
\sigma_t(\mathbf{x})=c\left(\frac{r_0}{R(s)}\right)^2\left(1-\frac{\rho^2}{R(s)^2}\right)^2 M(\mathbf{x}),
\label{eq:fogfield}
\end{equation}
and is zero outside. Here, $L$ is the plume length, $c$ a density scale, and $(r_0/R(s))^2$ represents dilution as the plume expands~\cite{arya1999air}. The cross-sectional profile is given by $(1-\rho^2/R(s)^2)^2$~\cite{wyvill1986data, muller2003particle}, and $M(\mathbf{x})$ is multi-scale fractal noise~\cite{lagae2010survey}. This yields heterogeneous fog with a smooth overall distribution and local density fluctuations.

The resulting extinction field is placed in the scene as a participating medium whose scattering is defined by a single-scattering albedo $\omega$ and the Henyey--Greenstein phase function~\cite{henyey1941diffuse}. A temporally impulsive point light is placed at the sensor, and the time-resolved response including multiple scattering in the medium is computed by path tracing with a single surface bounce. The response is convolved with an emitted pulse whose full width at half maximum matches that of the real laser, producing received intensity waveforms that contain both the attenuation of object reflections and fog backscatter.

\subsection{SPAD Receiver Model}
The rendered waveform represents light intensity, whereas a SPAD LiDAR outputs counts of detected photons. Our receiver model reproduces count saturation due to the inability to distinguish multiple photons, pile-up due to the suppression of subsequent detections during dead time, and tailing due to output holding. Let $N$ be the number of SPAD elements per pixel. The event that element $n$ detects a photon in bin $k$ is modeled as a Bernoulli variable $X_{n,k}\in\{0,1\}$ with probability~\cite{gupta2019photon}
\begin{equation}
a_k = 1-e^{-\lambda_k/N},
\label{eq:detect_prob}
\end{equation}
where $\lambda_k$ is the expected number of photons arriving at the pixel, and photon arrivals at each element follow independent Poisson distributions. Since $X_{n,k}$ is at most one even when multiple photons arrive, counts saturate for strong reflections. An element that detects a photon holds an output of one for $T_d$ bins and cannot detect new photons during that period, which reproduces pile-up and tailing. The count $C_k$ of the pixel at bin $k$ is the number of elements whose output is one.

\subsection{Difficulty Control and Annotation}
With the spatial distribution of the density field in Eq.~\eqref{eq:fogfield} fixed, fog density is controlled by the scale $c$. For rays hitting vehicles or pedestrians, we define the residual ratio as the mean ratio of the expected counts with and without fog at the peak position of the fog-free waveform. Targeting residual ratios of 0.80, 0.50, and 0.25 for Easy, Moderate, and Hard, respectively, $c$ is determined for each viewpoint by bisection on a logarithmic scale. Because this residual ratio does not account for the suppression of object reflections by fog backscatter, the final difficulty is evaluated on the generated data with the same procedure as Fog-FWL-Real (Sec.~\ref{sec:real}). Using the corresponding fog-free waveform of each frame, \textit{Object} and \textit{Fog} labels are assigned with the same procedure as for the real data, and the remaining bins are labeled \textit{Noise}. 3D bounding boxes are obtained from the scene layout.

\section{Dataset Statistics and Analysis}
\label{sec:stats}

\begin{table}[t]
\centering
\caption{Number of frames in Fog-FWL by difficulty level.}
\label{tab:stats}
\setlength{\tabcolsep}{5pt}
\begin{tabular}{lrrrr}
\toprule
Subset & Easy & Moderate & Hard & Total \\
\midrule
Fog-FWL-Real & 4,007 & 2,451 & 1,069 & 7,527 \\
Fog-FWL-Sim & 1,857 & 3,969 & 6,174 & 12,000 \\
\midrule
Total & 5,864 & 6,420 & 7,243 & 19,527 \\
\bottomrule
\end{tabular}
\end{table}

\begin{table}[t]
\centering
\caption{Ratio of object rays whose \textit{Object} label is included in the top-$k$ strongest peaks in Fog-FWL-Real.}
\label{tab:topk_accuracy_by_difficulty}
\setlength{\tabcolsep}{4pt}
\begin{tabular}{lrrrr}
\toprule
Difficulty & Frames & Top-1 & Top-2 & Top-3 \\
\midrule
Easy ($r \geq 0.70$) & 4,007 & 60.7\% & 79.7\% & 87.1\% \\
Moderate ($0.40 \leq r < 0.70$) & 2,451 & 26.4\% & 53.4\% & 65.8\% \\
Hard ($r < 0.40$) & 1,069 & 11.6\% & 25.5\% & 34.9\% \\
\bottomrule
\end{tabular}
\end{table}

Table~\ref{tab:stats} summarizes the number of frames per subset and difficulty level. Fog-FWL contains 19,527 frames in total. Fog-FWL-Real covers all three difficulty levels with more than 1,000 frames each, and Fog-FWL-Sim places more weight on the Moderate and Hard conditions, in which object reflections are strongly attenuated.

To quantify how much object information is lost by peak detection, Table~\ref{tab:topk_accuracy_by_difficulty} reports, for each difficulty level of Fog-FWL-Real, the ratio of rays inside 3D bounding boxes whose \textit{Object} label is included in the $k$ strongest peaks. Under Easy conditions, the object return is included in the top three peaks for 87.1\% of rays, whereas this ratio decreases to 65.8\% for Moderate and only 34.9\% for Hard. Even when using the strongest peak only, which corresponds to a single-return point cloud, the object return is captured for only 11.6\% of rays under Hard conditions. Consequently, point clouds based on peak detection lose most of the object surface in dense fog, and methods operating on such point clouds have limited observations to work with. These results show that recovering fog-hidden objects requires operating directly on FWL waveforms, and that the Hard split of Fog-FWL-Real provides a challenging benchmark for this task.

\section{Conclusion}
\label{sec:conclusion}

We presented Fog-FWL, an FWL dataset foundation for recovering objects hidden by fog. Fog-FWL-Real provides 7,527 real foggy FWL frames with histogram-level labels, 3D bounding boxes, and attenuation-based difficulty levels, and Fog-FWL-Sim provides 12,000 frames generated by a physics-based simulator that integrates heterogeneous fog, time-resolved light transport, and SPAD receiver effects. Our analysis shows that conventional peak detection discards most object returns in dense fog, motivating waveform-level recovery. The real-world data are currently limited to artificial fog and nearby vehicles, and extending the dataset to natural fog, more diverse objects, and other LiDAR sensors is left for future work. Fog-FWL will be publicly released upon acceptance. Benchmark results of recovery methods and downstream 3D object detection on Fog-FWL will be reported in a future version of this paper.

\bibliographystyle{IEEEtran}
\bibliography{main}

\end{document}